\PassOptionsToPackage{table}{xcolor}

\documentclass[10pt,twocolumn,letterpaper]{article}

\usepackage[pagenumbers]{wacv} 

\usepackage{algorithm}
\usepackage{algorithmic}
\usepackage{amsmath}
\usepackage{adjustbox}
\usepackage{graphicx}
\usepackage{multirow}
\usepackage{amssymb}
\usepackage{booktabs}
\usepackage{bm} 
\usepackage{ulem} 

\definecolor{wacvblue}{rgb}{0.21,0.49,0.74}
\usepackage[pagebackref,breaklinks,colorlinks,allcolors=wacvblue]{hyperref}

\def\wacvPaperID{902} 
\def\confName{WACV}
\def\confYear{2026}

\title{MVAT: Multi-View Aware Teacher for Weakly Supervised 3D Object Detection}

\author{
Saad Lahlali \and
Alexandre Fournier Montgieux \and
Nicolas Granger \and
Hervé Le Borgne \and
Quoc Cuong Pham
\\[2mm] 
Université Paris-Saclay, CEA, List, F-91120, Palaiseau, France \\
{\tt\small firstname.lastname@cea.fr}
}

\begin{document}

\maketitle
\begin{abstract}

Annotating 3D data remains a costly bottleneck for 3D object detection, motivating the development of weakly supervised annotation methods that rely on more accessible 2D box annotations. However, relying solely on 2D boxes introduces projection ambiguities since a single 2D box can correspond to multiple valid 3D poses. Furthermore, partial object visibility under a single viewpoint setting makes accurate 3D box estimation difficult.
We propose MVAT, a novel framework that leverages temporal multi-view present in sequential data to address these challenges.
Our approach aggregates object-centric point clouds across time to build 3D object representations as dense and complete as possible.
A Teacher-Student distillation paradigm is employed: The Teacher network learns from single viewpoints but targets are derived from temporally aggregated static objects. Then the Teacher generates high quality pseudo-labels that the Student learns to predict from a single viewpoint for both static and moving objects. The whole framework incorporates a multi-view 2D projection loss to enforce consistency between predicted 3D boxes and all available 2D annotations. Experiments on the nuScenes and Waymo Open datasets demonstrate that MVAT achieves state-of-the-art performance for weakly supervised 3D object detection, significantly narrowing the gap with fully supervised methods without requiring any 3D box annotations.
Our code is available in our public repository (\href{https://github.com/CEA-LIST/MVAT}{code}).
\end{abstract}

\section{Introduction} \label{sec:intro}
3D object detection is a fundamental perception task for applications like autonomous driving and robotics. One of the primary bottlenecks for this task remains the high cost of acquiring accurate 3D bounding box annotations for objects. Annotating a 3D bounding box takes on average 114 seconds \cite{song2015sun}, whereas 2D boxes take only 7-35 seconds \cite{papadopoulos2017extreme}. This potential 3-16x reduction in annotation time has motivated the development of weakly supervised methods that leverage cheaper 2D box annotations instead.

However, this cost-saving measure introduces a critical challenge: projection ambiguity.
A single 2D bounding box corresponds to an ambiguous space of potential 3D box locations, making it a significant challenge for a model to learn an object's true 3D shape and location.
To address this, current methods often rely on heuristics or priors to estimate the 3D box from image annotations~\cite{Lahlali_2025_WACV, zhang2024general}.

\begin{figure}[tb]
    \centering
    \includegraphics[width=1\linewidth]{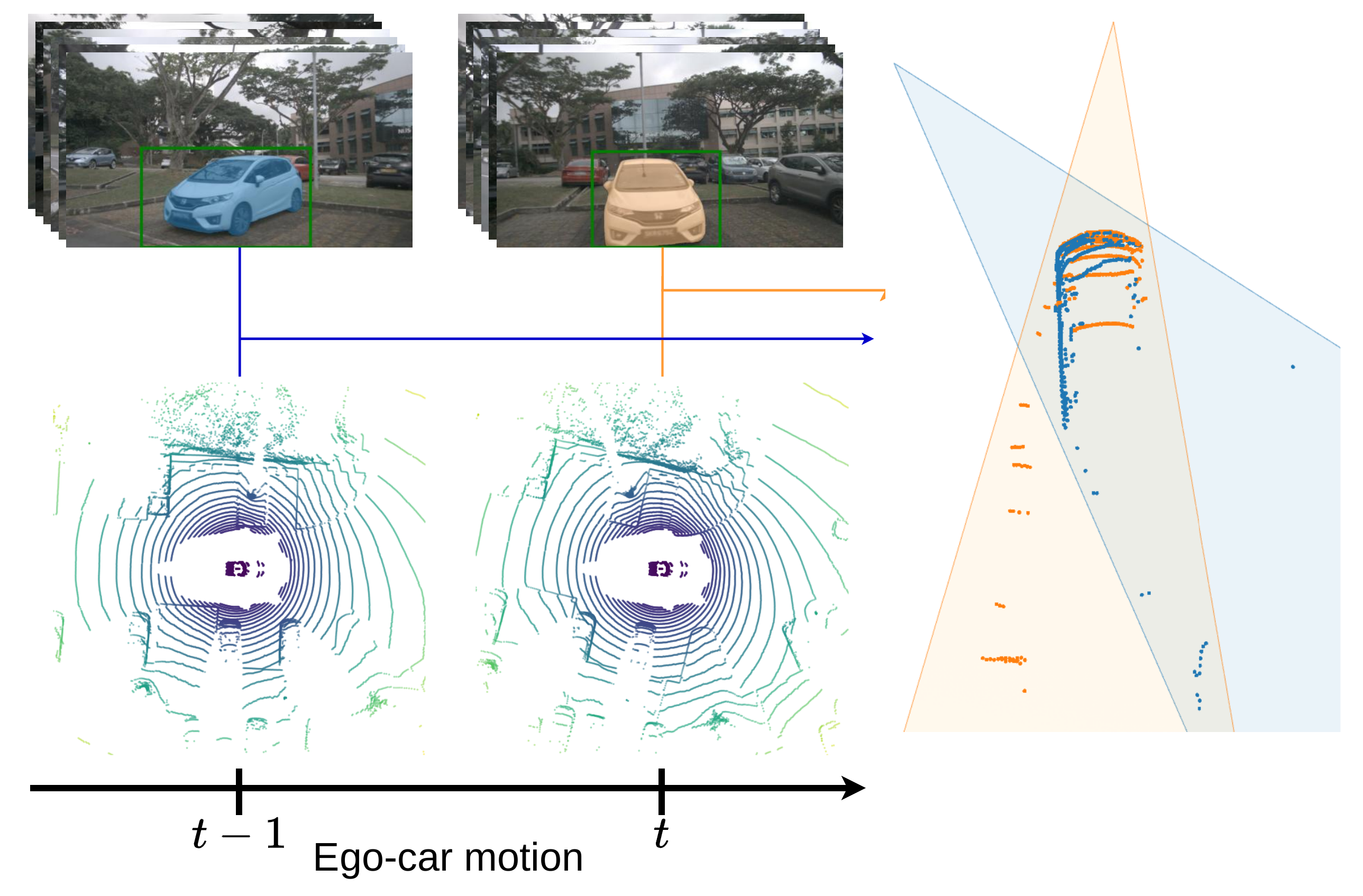}
    \caption{\textbf{Reducing Ambiguities with Temporal Multi-View.} The observation at time $t$ is a challenging case where the vehicle's length and center are difficult to infer from a single viewpoint without strong priors. By incorporating the view from time $t-1$, this ambiguity is lifted. The intersection of the viewing frustums from both time steps provides powerful geometric constraints, directly addressing the projection ambiguity challenge.}
    \label{fig:aggregation}
\end{figure}

These methods, however, overlook a powerful source of information that is naturally available in most real-world data-gathering scenarios.  In autonomous driving for example, data is captured continuously while the sensor is moving in the scene.
The ego-car’s continuous motion allows its sensors -- both cameras and LiDAR -- to capture multiple views of the same object over time, effectively creating a natural multi-view setup that we propose to exploit.
It is crucial to frame this within the objective of weakly supervised learning: building an automatic 3D annotator. Since this is an offline task performed once to label a dataset, real-time and causality constraints are not required and the annotation model can use all available frames.
Temporal constraints only apply to the final inference model trained with the annotated dataset.

\autoref{fig:aggregation} visually shows the benefits of our approach. Existing weakly supervised approaches try to annotate 3D boxes from a single view. However, it clearly appears that relying solely on the second viewpoint (observation at time $t$) is challenging since, without priors on the object's length, it is impossible to estimate its size and center precisely. Furthermore, the 2D box annotation cannot help disambiguate in this case since many 3D boxes can fit in the orange frustum shown on the right of the figure in bird eye's view. To solve this, existing approaches rely on category-specific shape and size priors.
Instead, we propose to simply combine several viewpoints of the same object to leverage all the data at our disposal, and thus not rely on any heuristic or prior.
In particular, just by adding the information available at time $t-1$, the ambiguity is largely reduced since the side of the car is now visible and the 3D box is naturally constrained since it has to be within the intersection of both frustums. The impact of leveraging viewpoints from multiple frames is twofold:
\begin{itemize}
    \item Several 3D views provide more 3D points which enriches the 3D representation of the object;
    \item Combining several 2D views constrains the possible 3D boxes since the different frustums, obtained from the 2D box annotations, intersect in a smaller volume.
\end{itemize}

Combining modalities and using all the information available over time allows us to get more precise 3D box estimations. This temporal multi-view consistency showed its advantages under the fully supervised setting \cite{park2022time, wang2023exploring}, yet it remains unexplored for weakly supervised learning. To our knowledge, no prior weakly supervised work has effectively exploited this temporal multi-view data to resolve these core challenges.

In this work, we introduce \textbf{MVAT (Multi-View Aware Teacher)}, a novel weakly supervised 3D detection framework designed to exploit temporal multi-view consistency based on a Teacher-Student distillation pipeline.
In order to provide 3D supervision to the Teacher network, we produce coarse box annotations for static objects by aggregating the LiDAR points over multiple frames. The model itself is trained using a single view as input in order to gain the ability to detect moving objects as well.
During the distillation phase, the Student network also takes as input one viewpoint for both static and moving objects, while the Teacher model is supplied with multi-view inputs for static objects in order to improve its reliability.
This strategy enables the Student to effectively learn the underlying 3D geometry and handle challenging cases like occlusions and moving objects where point aggregation is not directly applicable.
The training of both the Teacher and the Student is further refined by a multi-view projection 2D loss, which ensures consistency between the predicted 3D box and the 2D box ground truth annotations across all available views.

Our contributions can be summarized as follows:
\begin{enumerate}
\item We introduce MVAT, a novel Teacher-Student framework for weakly supervised 3D detection that is the first to effectively leverage temporal multi-view data to resolve projection ambiguity.
\item We propose a robust method for generating high-quality 3D object representations and pseudo-labels by aggregating sparse point clouds over time, guided by 2D annotations.
\item We employ a multi-view 2D projection loss that enforces geometric consistency across the sequence, serving as a powerful supervisory signal.
\item Our single-frame 3D annotator pipeline provides labels to train a 3D detector which achieves state-of-the-art performance in the weakly supervised paradigm on the nuScenes dataset, even for challenging cases involving moving or heavily occluded objects. Our pipeline shows consistent results on Waymo as well. To our knowledge, it is the first time this dataset is used in the weakly-supervised settings (2D boxes).
\end{enumerate}

\section{Related Works}
\label{sec:Related_Works}

\subsection{Weakly Supervised 3D Object Detection}
The weakly supervised setups aims to predict 3D bounding boxes using only 2D ground truth bounding box annotations, without access to any 3D ground truth.
The central challenge for weakly supervised 3D detection is resolving the inherent \textit{projection ambiguity} of a single 2D bounding box. The dominant strategy in the literature is to compensate for the lack of 3D information by introducing various forms of priors or external guidance, all while operating on a single frame.

Early methods relied heavily on class-specific geometric heuristics. For instance, FGR~\cite{wei2021fgr} and GAL~\cite{yin2023gal} craft explicit rules to estimate 3D box parameters for cars, which are then refined to align with 2D projections. More recent approaches have sought more generalizable priors. ALPI~\cite{Lahlali_2025_WACV} leverages class-average size priors to construct synthetic 3D proxy labels, while GGA~\cite{zhang2024general} uses Large Language Models to generate geometric priors. Others, like VG-W3D~\cite{huang2024weakly}, use the pseudo-labels generated by FGR as initialisation to guide their model.

\textit{Unlike existing weakly supervised methods that operate on a single frame and compensate for ambiguity using heuristics or class-specific priors, our approach directly addresses the ambiguity itself. We replace reliance on priors with geometric evidence gathered from multiple viewpoints over time, a source of information previously untapped in this setting.}

\subsection{Semi-weakly Supervised 3D Object Detection}
This alternative paradigm operates on a pragmatic compromise: it uses a small set of 3D box annotations to resolve the ambiguity in a much larger set of weakly-labeled data. This approach has primarily evolved along two branches:

The first branch uses 2D bounding boxes as weak label but requires a non-trivial fraction of the training data-often over 10\%-to be fully annotated with 3D boxes~\cite{MTA, CAT, MAP-Gen}. While these methods achieve strong results, their reliance on a substantial corpus of 3D ground truth presents a significant cost barrier to scalability and has typically constrained their application to a small number of object classes (\textit{Cars}).

The second branch aims to reduce the need for full 3D boxes by using "weaker", but still 3D-native, labels such as BEV centers~\cite{meng2021towards}, 3D centers~\cite{zhang2023simple}, or single point labels~\cite{gao2024leveraging}. This still raises a question of annotation costs since these weak 3D labels are inherently more expensive than annotating 2D boxes. They all necessitate time-consuming interaction with a 3D point cloud viewer, as noted by Meng et al.~\cite{meng2021towards}: Locating a BEV center often requires first drawing a 2D box to isolate the object, making the process inherently more complex than using image-based 2D boxes alone. Furthermore, these labels create an information gap; a single 3D point provides no signal for an object's size or orientation, forcing the model to rely entirely on the small fully-annotated subset to learn these crucial properties.

\textit{We distinguish our work from the semi-weakly supervised paradigm by adhering to a stricter, purely 2D-supervised problem. We require no fraction of 3D box annotations, nor do we depend on weak 3D labels like BEV centers or 3D points, which still carry a significant annotation cost. Our method relies exclusively on the most scalable form of supervision: 2D image boxes.}

\subsection{Temporal Fusion in 3D Object Detection}

While the weakly supervised field has focused on single-frame solutions, research in fully- and semi-supervised settings has long established the power of temporal fusion. By aggregating information across multiple frames, these methods achieve significant performance gains, particularly in handling occlusion and detecting distant objects.
For example, approaches like Time Will Tell~\cite{park2022time} demonstrate that long-term temporal integration is key to robust detection. Others, such as StreamPETR~\cite{wang2023exploring} and HOP~\cite{zong2023temporal}, develop efficient object-centric frameworks to model object states and motion cues over time, proving that a detector's understanding of spatial semantics is deeply enhanced by temporal context.

\textit{The success of temporal fusion in fully-supervised detection is well-documented. However, a clear dichotomy exists in the literature: its benefits have not been explored for the fundamental challenges of the weakly supervised problem. Our work is the first to bridge this divide. We demonstrate that the multi-view consistency inherent in sequential data is not just a performance booster for well-supervised models, but a powerful tool to bootstrap 3D understanding from 2D annotations alone.}

\section{Method}
\label{sec:method}

\begin{figure*}[!htbp]
  \centering
   \includegraphics[width=\linewidth]{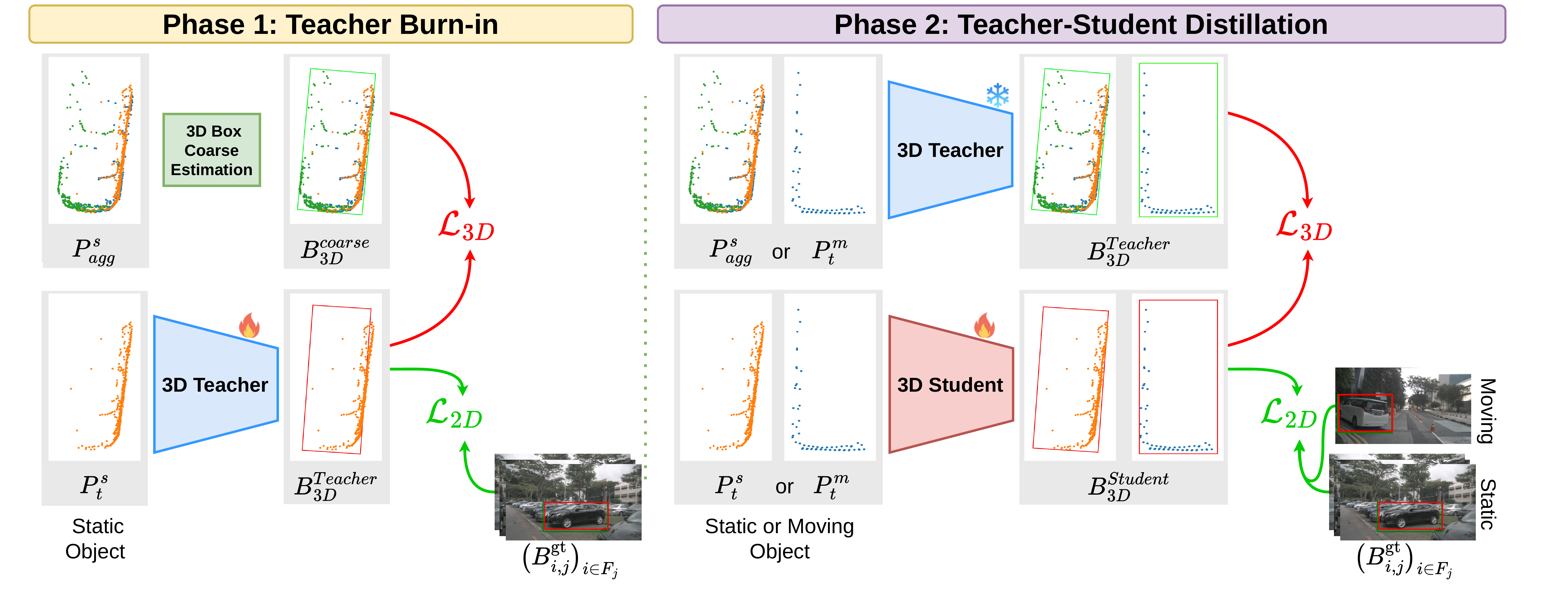}
   \caption{\textbf{Overview of our Two-Phase Teacher-Student Pipeline.} Our method bootstraps 3D understanding from only 2D annotations by training a Teacher and then distilling its knowledge into a  Student. \textbf{Phase 1: Teacher Burn-in.} A 3D Detector Teacher is trained to infer 3D boxes from single-frame views $P_t^s$, using its own native 3D loss $\mathcal{L}_{3D}$, ensuring compatibility across different 3D detection frameworks. To enable this, we first generate 3D Box Coarse Estimation pseudo-labels (\autoref{subsec:3d_box_coarse_estimation}) by aggregating temporal point clouds of static objects $P_{\text{agg}}^{s}$ (\autoref{subsec:temporal_aggregation_data_representation}). The Teacher is then supervised by these coarse labels alongside a corrective multi-view 2D projection loss $\mathcal{L}_{2D}$ (\autoref{subsec:Teacher_training_multi_view_supervision}). \textbf{Phase 2: Teacher-Student Distillation.} The Teacher then generates high-quality pseudo-labels for aggregated point clouds for static objects ($P_{\text{agg}}^{s}$) and isolated point clouds for moving objects ($P_{t}^{m}$). A 3D Detector Student learns from these labels under single-frame conditions, alongside a corrective multi-view 2D projection loss (\autoref{subsec:Teacher_Student_distillation}). This phase allows the Student to better generalise to both moving and static objects. }\label{fig:overview}
\end{figure*}

Our objective is to devise a 3D annotator framework that learns to annotate accurate 3D detections from sequences of images and point cloud pairs,
\textbf{annotated solely with 2D bounding boxes}. The central challenge is the inherent geometric ambiguity of projecting a 3D world onto a 2D plane. Our method, the Multi-View Aware Teacher (MVAT), is founded on a key insight: while a single viewpoint is ambiguous, sequential data from a moving ego-vehicle naturally provides multiple views of static objects, offering powerful geometric constraints to resolve this ambiguity.

MVAT leverages this temporal multi-view consistency within a two-phase Teacher-Student distillation framework, as illustrated in \autoref{fig:overview}. The core strategy is to first train a Teacher network on high-quality, aggregated point cloud representations of static objects to learn robust 3D geometry. This Teacher then generates robust 3D box pseudo-labels for the entire dataset (both static and moving objects), which a Student network learns to predict from single-frame input. This distillation process creates a powerful 3D annotator, capable of handling hard single-frame conditions. The entire framework is architecture-agnostic, allowing for any off-the-shelf 3D detector to be used for the Teacher and Student.


\subsection{Object-Centric Point Cloud Extraction}
\label{subsec:object_centric_pc_extraction}

The first step of our method is to isolate the points corresponding to each object instance in every frame. Given a sequence of sensor data $\{(I_t, P_t^{\text{raw}})\}_{t=1}^T$ and their associated ground truth 2D bounding boxes $B_{t,j}^{\text{gt}}$ for each object $j$ at time $t$, we first generate a segmentation mask. We denote by \(\mathcal{F}_j\) the set of image frames in which the object \(j\) appears, with each frame \(I_i\) (for \(i \in \mathcal{F}_j\)) annotated with the corresponding ground truth 2D bounding box \(B_{i,j}^{\text{gt}}\). We obtain a segmentation mask \(M_t\) via SAM~2 \cite{ravi2024sam} using \(I_t\) and  \(B_{i,j}^{\text{gt}}\) as a box prompt.
We then filter the raw point cloud by projecting each point onto the image plane and retaining only those that fall within the mask. This yields an object-centric point cloud for each instance in each frame:
\begin{equation}
    P_{t,j} = \{\, p \in P_t^{\text{raw}} \mid \pi(p) \in M_{t,j} \,\}
    \label{eq:point_extraction}
\end{equation}
where $\pi(\cdot)$ is the projection function. This strategy effectively isolates relevant foreground objects, preparing them for the temporal analysis that follows.

\subsection{Temporal Aggregation and Data Representation}
\label{subsec:temporal_aggregation_data_representation}

To overcome the sparsity and ambiguity of a single view, our key strategy is to aggregate point clouds over time. However, aligning points for moving objects without ground-truth motion information is arduous. We therefore restrict aggregation to \textbf{static instances}, which constitute a significant majority of objects in typical driving scenarios. Analysis of the nuScenes dataset shows that approximately 74\% of objects instances are static, and no class is composed exclusively of moving objects. We provide a detailed analysis of the per-class temporal visibility in the supplementary material, which further motivates our approach.

We identify static objects by analyzing the temporal consistency of their point cloud centroids across frames, more details about this step are provided in the supplementary materials. This allows us to define three distinct sets of point cloud inputs that are fundamental to our Teacher-Student pipeline:
\begin{itemize}
    \item \textbf{Aggregated Static Points ($P_{\text{agg}}^{s}$):} For each static object, we aggregate its point clouds from the extraction step ($P_{\text{agg}}^{s} = \bigcup_{t \in \mathcal{F}_j} P_{t,j}^{s}$). This creates a dense, geometrically complete representation used to generate high-quality 3D pseudo-labels.
    \item \textbf{Isolated Static Points ($P_{t}^{s}$):} The sparse, partial point cloud of a static object as seen from only a single frame $t$.
    \item \textbf{Isolated Moving Points ($P_{t}^{m}$):} The sparse, partial point cloud of a moving object from a single frame $t$.
\end{itemize}

This data separation is motivated by a following insight: from the perspective of a single timestamp $t$, an object's point cloud appearance is \textbf{independent of its motion state}. The learning challenge presented by an isolated static view ($P_{t}^{s}$) is virtually identical to that of an isolated moving view ($P_{t}^{m}$).
This key insight enables our distillation strategy: training the Teacher on isolated static views creates representations that naturally generalize to moving objects, allowing it to provide high-quality pseudo-labels for both cases that the Student learns to refine.




\subsection{3D Box Coarse Estimation}
\label{subsec:3d_box_coarse_estimation}

The goal of this step is to derive a coarse-yet-robust 3D pseudo-box, $B_{\text{3D}}^{\text{coarse}}$, from each aggregated static point cloud, $P_{\text{agg}}^{s}$. As seen in \autoref{fig:aggregation}, even when generating accurate 2D masks using SAM 2, noisy points remain in cropped 3D point cloud $P_{t,j}$. This is mainly due to imperfect alignment between camera and LiDAR. A cleaning process is needed to remove such noise that can severely impact the quality of the 3D coarse boxes' estimation. We first cluster $P_{\text{agg}}^{s}$ using DBSCAN~\cite{ester1996density} and select the dominant point cluster, $C^*$. To ensure the quality of our initial labels, we then filter out any instance where $C^*$ is too sparse or was constructed from an insufficient number of temporal views. This guarantees that only high-quality aggregated instances are used to bootstrap the learning process.

From the clean cluster $C^*$, we estimate the box parameters. First, by applying Principal Component Analysis (PCA) \cite{mackiewicz1993principal} to $C^*$ in the bird’s-eye view (BEV, i.e., the horizontal \(xy\) plane) to determine the object's 3D box parameters in BEV (visualizations are provided in supplementary material). PCA yields principal axes \(\bm{v_1}\) and \(\bm{v_2}\) from which we compute center \((c_x, c_y)\) as the midpoints of the min/max points along each principal axis, size \((l, w)\) as the differences between the maximum and minimum values along these axes, and heading \(\theta\) as the orientation of \(\bm{v_1}\). Then, the center and size along the vertical axis are estimated separately using the points in \(C^*\): the \(z\)-axis center \(c_z\) is the average of the minimum and maximum \(z\) values, and the height \(h\) is their difference.

The resulting $B_{\text{3D}}^{\text{coarse}} = \bigl( c_x, c_y, c_z, l, w, h, \theta \bigr)$ is not expected to be perfectly aligned; rather, it serves as a coarse, data-driven initial hypothesis that bootstraps the Teacher's learning process.

\subsection{Teacher Burn-in}
\label{subsec:Teacher_training_multi_view_supervision}

The goal of the Teacher network is to develop a deep understanding of 3D geometry. During a burn-in phase, we train an off-the-shelf 3D detector exclusively on the object samples kept during the previous step.\\
\textbf{3D supervision loss} ($\mathcal{L}_{\text{3D}}$). Notably, the network takes as input the isolated points of static objects $P_{t}^{s}$. This partial-view training regime is central to our methodology. 
By training the network to predict the object's 3D box parameters $B_{\text{3D}}^{\text{coarse}}$ from sparse, single-frame point clouds, we compel it to achieve robust predictions to both static and moving objects. 
Through its native 3D loss $\mathcal{L}_{\text{3D}}$, the Teacher learns from these challenging partial views, which creates a foundation for the subsequent knowledge transfer to the Student network, which handles real-world single-frame inference conditions.\\
\textbf{2D grounded supervision loss} ($\mathcal{L}_{\text{2D}}$). This loss provides a powerful supervision using all available 2D box annotations of the object. Specifically, we project each predicted 3D box $B_{\text{3D}}^{\text{Teacher}}$ of the object $j$ into every camera view $i \in \mathcal{F}_j$ where it is visible to produce the predicted 2D box $B_{i,j}^{\text{pred}}$. The per-object loss is then the average Generalized Intersection-over-Union (GIoU) loss between the projected 2D predictions $B_{i,j}^{\text{pred}}$ and the ground-truth 2D boxes $B_{i,j}^{\text{gt}}$:
\begin{equation}
\mathcal{L}_{2D}^{(j)} = \frac{1}{|\mathcal{F}_j|} \sum_{i \in \mathcal{F}_j} \Bigl( 1 - \operatorname{GIoU}\bigl(B_{i,j}^{\text{pred}}, B_{i,j}^{\text{gt}}\bigr) \Bigr)
\end{equation}
During training, we average these per-object losses to compute the final $\mathcal{L}_{2D}$ for the batch. This ensures each object contributes equally to the gradient, preventing instances that are visible in more frames from dominating the training signal.\\
The total training loss for the Teacher is the following weighted sum:
\begin{equation}
\mathcal{L}_{\text{Teacher}} =  \mathcal{L}_{\text{3D}} + \lambda \mathcal{L}_{2D}
\label{eq:Teacher_loss}
\end{equation}

\subsection{Teacher-Student Distillation}
\label{subsec:Teacher_Student_distillation}

The Teacher network is trained on only a filtered subset of static instances and has therefore not seen the full data distribution. The final stage aims to distill its specialized knowledge into a Student network. The goal is to create a robust, generalist detector that is trained on all object instances (static and moving) and can handle the challenges of real-world, single-frame inference.\\
\textbf{Teacher (Pseudo-labeling):} To generate high-fidelity pseudo-labels, the Teacher receives the aggregated point clouds ($P_{\text{agg}}^{s}$) for static objects and keeps using isolated point clouds ($P_{t}^{m}$) for moving objects. In essence, we train the Teacher on a difficult task to make it powerfull, and then we give it an easier task to generate the best possible answers. This strategy is precisely what allows our framework to create high-quality pseudo-labels. We filter out object instances wrongly classified by the Teacher using the ground truth annotation (supplementary materials).\\
\textbf{Student (Training):} The Student is then trained to replicate the Teacher's prediction but under more challenging, real-world conditions. It exclusively receives single-frame inputs ($P_{t}^{s}$ for static objects and $P_{t}^{m}$ for moving ones) drawn from the entire dataset (with the exception of filtered object during the previous pseudo labeling phase). \\
Similarly to the Teacher training loss, the total loss is the weighted sum of the 2D multi-view Loss and the 3D loss between the teacher pseudo label \(B_{\text{3D}}^{\text{Teacher}}\) and the Student prediction \(B_{\text{3D}}^{\text{student}}\):

\begin{equation}
\mathcal{L}_{\text{Student}} =  \mathcal{L}_{\text{3D}} + \gamma \mathcal{L}_{2D}
\label{eq:Student_loss}
\end{equation}

\noindent Using this dual supervision for the Student training as well ensures that the Student too develops a more accurate and robust 3D box estimation, especially in the case of incorrect pseudo labeling.


\section{Experiments}
\label{sec:experiments}

\subsection{Datasets and Evaluation Metrics}
\begin{table*}[tb]
\centering
 \begin{adjustbox}{width=\linewidth}
\begin{tabular}{ccccccccccccccc}
\toprule
\textbf{Paradigm} & \textbf{3D boxes} & \textbf{3D Annotator} & \textbf{mAP} & \textbf{SPNDS} & \textbf{Car} & \textbf{Truck} & \textbf{C.V.} & \textbf{Bus} & \textbf{Trailer} & \textbf{Barrier} & \textbf{Motor.} & \textbf{Bike} & \textbf{Ped.} & \textbf{T.C.}
\\
\cmidrule(){1-15}
\rowcolor{gray!20}
fully & 100\% & \textit{Oracle}  & 58.8 & 61.80 & 84.8 & 57.5 & 18.3 & 69.3 & 34.8 & 68.5 & 57.2 & 42.8 & 85.3 & 69.9 \\
\cmidrule(){1-15}
semi  & 7\% & \cite{wang2021semi} & 36.5 & - & 74.9&  29.9& 3.8& 31.3& 10.1 &45.6 & 31.9 & 12.3& 75.6 & 49.1 \\
\cmidrule(){1-15}

\multirow{4}{*}{semi-weakly}   & \multirow{2}{*}{5\%} & Point-DETR3D \cite{gao2024leveraging} & 50.80 &  53.99 & -&  -& -& -& - &-  & - & -& - & - \\ 

  &  & \textbf{MVAT (ours)} & 54.1 & 56.4 & 83.0 & 50.2 & 17.0 & 61.4 & 31.2 & 61.8 & 53.4 & 39.8 & 82.8 & 60.0\\
  
  \cmidrule(){2-15}
  
& \multirow{2}{*}{2\%} & Point-DETR3D \cite{gao2024leveraging} & 47.49 & 48.21 & -&  -& -& -& - &-  & - & -& - & - \\

  &  & \textbf{MVAT (ours)} & 49.5 & 51.7 & 82.7 & 42.8 & 15.7 & 54.2 & 29.1 & 55.0 & 49.4 & 33.9 & 79.0 & 53.2\\

\cmidrule(){1-15}
 \multirow{3}{*}{weakly } &  \multirow{3}{*}{0\%} & ALPI~\cite{Lahlali_2025_WACV}   & 41.8 & - & 81.0& 33.9 & 7.7 & 45.3 & 30.0 & 40.0 & 42.7 & 26.3 & 73.9 & 37.0 \\ 
  &  & \textbf{MVAT (ours)} & \textbf{47.6} & \textbf{49.1} & \textbf{82.3} & \textbf{41.1} & \textbf{15.4} & \textbf{50.7} & \textbf{32.2} & \textbf{55.8} & \textbf{47.2} & \textbf{28.5} & \textbf{76.9} & \textbf{45.6}  \\ 
  
& & improvement  & {\color{teal}+5.8} & - & {\color{teal}+1.3} & {\color{teal}+7.2} & {\color{teal}+7.7} & {\color{teal}+5.4} & {\color{teal}+2.2} & {\color{teal}+15.8} & {\color{teal}+4.5} & {\color{teal}+2.2} & {\color{teal}+3.0} & {\color{teal}+8.6} \\
\bottomrule
\end{tabular}
\end{adjustbox}
\caption{3D object detection performance on the nuScenes validation set. We report results using the mAP and SPNDS metrics. ‘C.V.’, ‘Ped.’, and ‘T.C.’ are short for 'construction vehicle', 'pedestrian', and 'traffic cone', respectively. CenterPoint~\cite{yin2021center} is used as the 3D detector. Point-DETR3D \cite{gao2024leveraging} doesn't provide mAP scores for each class. We bold the best overall results and list the relative improvements based on the 3D annotator ALPI~\cite{Lahlali_2025_WACV} for better illustration.}
\label{tab:eval_nuscene}
\end{table*}
\textbf{nuScenes Dataset.}  To evaluate our framework's ability to leverage temporal and multi-view information, we use the large-scale nuScenes dataset~\cite{nuscenes}. It consists of 1,000 driving sequences, divided into 700 for training and 150 for validation. For evaluation, we report the mean Average Precision (mAP) and the Static Properties nuScenes Detection Score (SPNDS) following \cite{gao2024leveraging}. The mAP is calculated over Bird's-Eye View (BEV) center distance thresholds. The SPNDS is derived from the standard nuScenes Detection Score (NDS) by removing velocity and attribute related measurements as methods in weakly supervised 3D object detection focus on the static properties of 3D bounding boxes. \\
\textbf{Waymo Open Dataset.} We also conduct experiments on the Waymo Open Dataset. We follow the official evaluation protocol, reporting the Average Precision (AP) for a 3D IoU threshold of 0.7 for vehicles and 0.5 for pedestrians and cyclists. Performance is evaluated across two difficulty levels: Level 1 (L1) for objects with more than 5 points, and Level 2 (L2), which includes objects with 5 or fewer points.

\subsection{Implementation Details}

Our 3D annotator is adapted from the PV-RCNN framework~\cite{shi2020pv}, \textbf{$\mathcal{L}_{\text{3D}}$} is detailed in the supplementary material. Since the annotator processes one object at a time, we configure the Region Proposal Network (RPN) to generate a maximum of 100 proposals, which are refined to approximately 20 final detections per object instance. Both Teacher and Student networks are trained from scratch. For both the nuScenes and Waymo datasets, the Teacher and Student models are each trained for 80 epochs with a global batch size of 64 on four NVIDIA A100 GPUs. 
To evaluate the pseudo-labels generated by our annotator, we train the widely-used CenterPoint~\cite{yin2021center} detector in its standard configuration. Specifically, we first use our converged Student network to re-annotate the entire training dataset with 3D pseudo-labels. The final performance is then reported on the CenterPoint model trained from scratch on this newly generated dataset. 
A detailed discussion of our experimental setup is available in the supplementary material.

\subsection{Comparisons with the State of the Art}

\textbf{Performance on nuScenes.}
As presented in \autoref{tab:eval_nuscene}, our method, MVAT, establishes a new state-of-the-art for weakly supervised 3D object detection on the nuScenes validation set. We achieve an mAP of 47.6\% and an NDS of 49.1\%, significantly outperforming the previous leading method, ALPI~\cite{Lahlali_2025_WACV}, by a margin of +5.8 mAP. Our weakly supervised approach now reaches 81.0\% of the performance of the fully supervised oracle (47.6 vs 58.8 mAP), demonstrating a significant step towards bridging the gap while using only 2D annotations.

The strength of our temporal, multi-view approach is particularly evident when analyzing performance on challenging object categories. For instance, long objects are frequently occluded, such as Truck (+7.2) and Bus (+5.4), aggregating views thus provides a complete understanding of the object's extent that is unavailable from a single frame. Similarly, performance soars for geometrically sparse objects like Barrier (+15.8) and Traffic Cone (+8.6). For these classes, our temporal aggregation builds a coherent and dense representation, directly validating our core hypothesis, our method narrowed the gap to the fully supervised upper bound to just 2.5 points. This suggests that while prior-based methods are effective, our approach's true strength lies in its ability to resolve geometric ambiguity without relying on heuristics and priors.

\noindent \textbf{Analysis in the Semi-Weakly Supervised Setting.}
To contextualize the performance of our purely 2D-supervised approach, we also evaluate MVAT in a semi-weakly supervised setting. To do this, we adapt our method to leverage a small fraction of 3D ground-truth boxes (2\% and 5\%) by using them as an additional, high-quality supervisory signal for the Teacher network during its burn-in phase. These ground-truth boxes are used alongside the coarse pseudo-labels derived from our temporal aggregation, effectively enriching the Teacher's training signal. 

We compare (\autoref{tab:eval_nuscene}) this setup against Point-DETR3D~\cite{gao2024leveraging}, a state-of-the-art method that uses a stronger weak label (3D point annotations) with the same percentage of 3D boxes. 
Our method with \textbf{zero} 3D box annotations (47.6 mAP) outperforms Point-DETR3D when it is given 2\% of the ground-truth 3D boxes (47.5 mAP). 
This validates that our multi-view temporal strategy is more informative than 3D point labels combined with a small amount of strong supervision.
At the 2\% setting, MVAT shows an improvement of +2.0 mAP and +3.5 SPNDS. This trend continues at the 5\% setting, where our method achieves advantages of +3.3 mAP and +2.4 SPNDS. The consistent mAP and SPNDS results suggest our approach is effective at accurately estimating the geometric properties of objects. We also note that with only 5\% of the 3D annotations, MVAT reaches 92\% of the performance of the fully supervised oracle (54.1 vs 58.8 mAP).
This trend suggests that our method is also more effective at leveraging strong supervision when it becomes available, presumably because our temporal aggregation provides a better geometric foundation for the model to refine.

\textbf{Performance on Waymo Open Dataset.}
To validate the robustness and scalability of our approach, we also evaluate MVAT on the Waymo Open Dataset as shown in \autoref{tab:table_waymo},.
To our knowledge, we are the first method to report performance metrics on this dataset in the weakly supervised paradigm.
For the L1 difficulty, our method achieves 91.3\% of the oracle's AP for Vehicles (72.1 vs. 79.0 AP), 89.4\% for Pedestrians (68.5 vs. 76.6 AP), and 87.4\% for Cyclists (59.6 vs. 68.2 AP). These results were obtained without the need to adapt hyper-parameters to the characteristics of this dataset, thus demonstrating the generality of our approach.

\begin{table}[tb]
\centering
\resizebox{\linewidth}{!}{
\begin{tabular}{cccccccc}
\hline
\multirow{2}{*}{\textbf{3D boxes}} & \multirow{2}{*}{\textbf{Annotator}} & \multicolumn{2}{c}{\textbf{Vehicle}} & \multicolumn{2}{c}{\textbf{Pedestrian}} & \multicolumn{2}{c}{\textbf{Cyclist}} \\
\cline{3-8}
& & L1  & L2  & L1  & L2  & L1  & L2  \\
\hline
\rowcolor{gray!20}
100\% & \textit{Oracle} & 79.0 & 79.0 & 76.6 & 72.3 & 68.2 & 66.2 \\
0\% & \textbf{MVAT (ours)} & \textbf{72.1} & \textbf{70.9} & \textbf{68.5} & \textbf{58.8} & \textbf{59.6} & \textbf{55.3} \\
\hline
\end{tabular}
}
\caption{3D object detection AP on the Waymo validation set using CenterPoint as the 3D detector. MVAT achieves performance relatively close to the fully supervised Oracle, similarly to the results obtained on Table 1,  across both Level 1 and Level 2 difficulty settings.}
\label{tab:table_waymo}
\end{table}

\begin{figure*}[tb]
  \centering
        \includegraphics[width=1\linewidth]{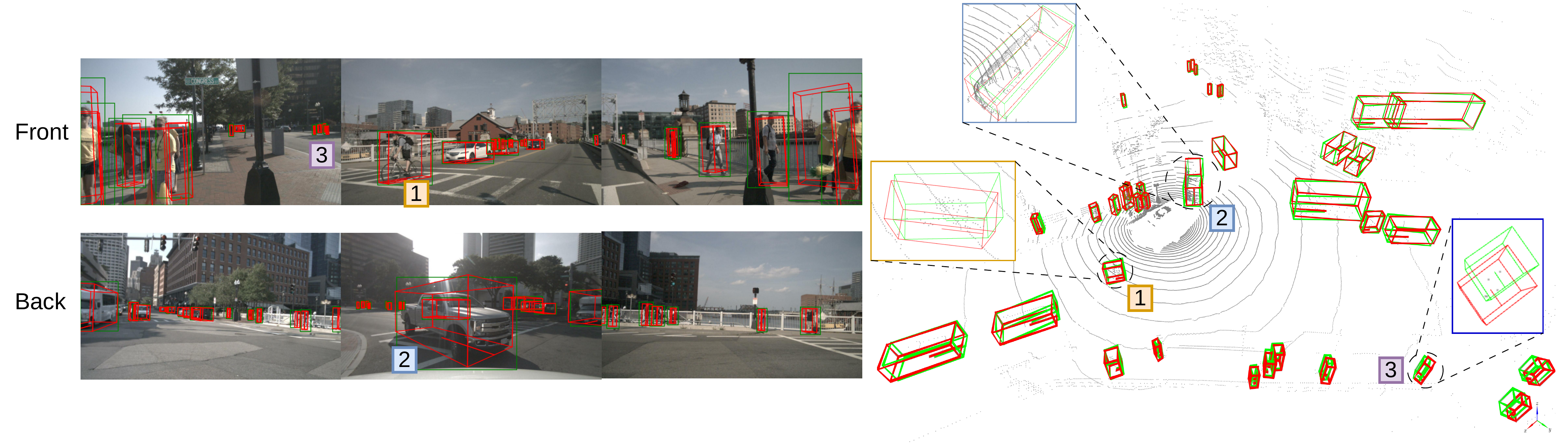}
    \caption{\textbf{Qualitative Results on a scene from nuScenes.} We visualize predictions generated by CenterPoint~\cite{yin2021center} trained with annotations from our MVAT annotator. The right panel displays a 3D view of the LiDAR point cloud. Ground-truth 3D boxes are shown in green and predictions are in red. The left one shows the six camera images with the projected 3D box predictions in red and the ground truth 2D box annotations (green) that served as MVAT's only ground truth during training.  Numbered insets provide a closer look at individual objects, including a \textit{Bicycle} (1), a \textit{Truck} (2), and a \textit{Pedestrian} (3). }
    \label{fig:visu}
\end{figure*}

\subsection{Ablation Studies}

\begin{figure}[tb]
    \centering
    \includegraphics[width=0.8\linewidth]{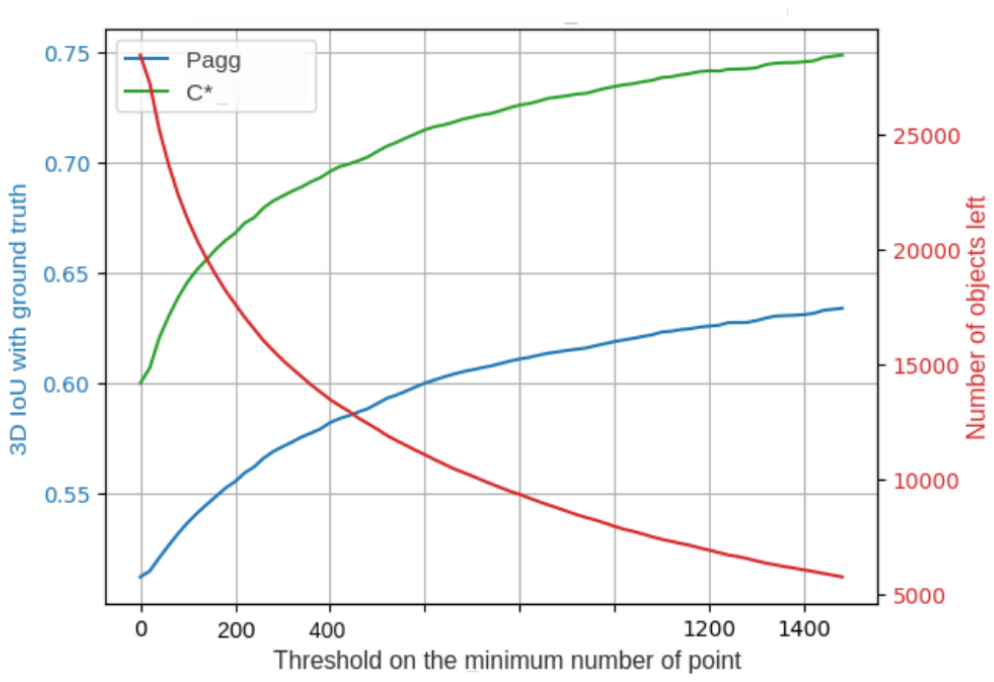}
    \caption{\textbf{Evaluation of the coarse 3D instance segmentation} without ($P_{agg}$) and with ($C^*$) DBSCAN clustering.}
    \label{fig:segmentation}
\end{figure}

\textbf{Quality of the 3D coarse segmentation.}
To assess the quality of our 3D segmentation pipeline explained in \autoref{subsec:object_centric_pc_extraction} and  \autoref{subsec:3d_box_coarse_estimation}, we evaluate the aggregated point cloud \(P_{\text{agg}}\) (obtained using SAM~2 masks) and the refined dominant cluster \(C^*\) (obtained via DBSCAN clustering) against the provided 3D instance segmentation ground truth generated from the 3D box ground truth. Our analysis focuses on the effect of imposing a minimum point threshold on \(C^*\) to filter out unreliable objects. \autoref{fig:segmentation} plots the average 3D IoU as a function of the minimum point threshold, for all classes of nuScene except \textit{traffic cone} and \textit{barrier} that are always static. With only SAM~2 segmentation, the average IoU is 0.53. After clustering, the IoU improves to approximately 0.60. Further filtering by increasing the minimum point threshold yields a substantial IoU gain as the threshold approaches 200 points.
However, this improvement comes at the cost of dramatically reducing the number of objects retained in the dataset.
In our experiments, we find that simply setting a minimum of 10 points for all classes is sufficient to filter out boxes that could impede the training of the Teacher model.

\begin{table}[tb]
\centering
\resizebox{0.75\linewidth}{!}{
\begin{tabular}{ccccc}
\toprule
\textbf{Class} & \textbf{Car} & \textbf{Pedestrian} & \textbf{Bicycle} & \textbf{Average}
\\
\cmidrule(){1-5}
3D IoU & 53.3 & 54.6 & 46.2 & 49.2 \\
\bottomrule
\end{tabular}}
\caption{\textbf{Quality of the 3D coarse estimations. }Evaluation in 3D IoU on nuScenes between the 3D coarse estimations \(B_{\text{3D}}^{\text{coarse}}\) and ground truth 3D boxes for the samples used to train the Teacher network. Additional quantitative results can be found in the supplementary materials.}
\label{tab:iou}
\end{table}

\noindent\textbf{Quality of the 3D coarse regression \(B_{\text{3D}}^{\text{coarse}}\).}
A fundamental component of our weakly supervised framework is the estimation of coarse 3D boxes that guide the Teacher network initially.
We evaluate the quality of these boxes using 3D IoU, which directly corresponds to the criteria used in PV-RCNN to classify positive and negative proposals. In \autoref{tab:iou} the 3D IoU results are reported for some classes of nuScenes. With an average 3D IoU of 49.2 across all categories, our coarse estimations provide significantly better initialization than random proposals, despite using only parameter-free algorithms (DBSCAN clustering and PCA) and no 3D annotations.
The highest IoU values are observed for categories with more consistent shapes and sizes (Pedestrian: 0.54, Car: 0.53), while more complex categories with higher shape variance (Bus: 43, Barrier: 45) present greater challenges.
Notably, these coarse estimations are of sufficient quality to be used as-is for training the teacher.

\noindent\textbf{Static-to-Moving Generalization.}
We conduct ablation studies to validate the effectiveness of our Teacher-Student distillation strategy for generalizing from static to moving objects. All results are presented in \autoref{tab:network_config}.
First, to quantify the value of \textit{temporal aggregation}, we compare the Teacher's performance on static objects. Using the complete, aggregated point cloud (Static, All) boosts performance to 67.8 AP, a significant +5.3 AP gain over using only a single frame (62.5 AP). This confirms that aggregation can contribute to produce better pseudo-labels. Next, we analyze the impact of our \textit{knowledge distillation} strategy. The Student network significantly outperforms its Teacher on identical single-frame inputs, improving performance on static objects from 62.5 to 67.4 AP (+4.9) and on moving objects from 62.1 to 66.7 AP (+4.6). This demonstrates that distillation successfully transfers the Teacher's expert knowledge into a more powerful and generalist final model.
The Student's near-equal performance on static (67.4 AP) and moving (66.7 AP) objects further validates that our approach does not introduce a domain gap between static and moving objects.

\begin{table}[tb]
 \centering
 \resizebox{0.7\linewidth}{!}{
 \begin{tabular}{cccc}
  \toprule
  \textbf{Network} & \textbf{Object Type} & \textbf{\# Frames} & \textbf{AP} \\
  \midrule
  \multirow{4}{*}{\textbf{Teacher}} & \cellcolor{gray!20} Static & \cellcolor{gray!20} Single & \cellcolor{gray!20} 62.5 \\
  & Static & All & 67.8 \\
  & Moving & Single & 62.1 \\
  \cmidrule{2-4}
  & All & All & 64.7 \\
  \midrule
  \multirow{4}{*}{\textbf{Student}} & Static & Single & 67.4 \\
  & Moving & Single & 66.7 \\
  \cmidrule{2-4}
  & All & Single & 67.0 \\
  \bottomrule
 \end{tabular}
 }
 \caption{\textbf{Performance on moving/static objects} depending on available input frames (nuScenes, all classes except 'Traffic Cone', 'Barrier'). Teacher uses aggregated frames for static, single for moving during pseudo-labeling.}
 \label{tab:network_config}
\end{table}

\noindent\textbf{Importance 3D Geometric Guidance.}
Relying on a multi-view 2D loss alone causes instability during training, as a few geometrically ambiguous cases can cause sharp spikes in the training gradient and hinder convergence. We found that our model fails to converge without any initial 3D guidance, which confirms previous findings from ALPI~\cite{Lahlali_2025_WACV} on the need for 3D supervision to stabilize training.

\section{Limitations and Perspectives}

MVAT's reliance on separating static and moving objects introduces pipeline complexity and may limit performance in dynamic environments like highways. Future work could follow two promising directions. \\
First, incorporating explicit motion modeling could improve aggregation quality for moving objects, potentially eliminating the need for static/moving separation and improving the model for non-rigid classes such as pedestrians.\\
Second, the training objective for both Teacher and Student networks could be enriched by introducing an auxiliary task to predict the spatial distribution of points from the aggregated 3D object views given only a single-frame input. This would provide additional geometric supervision beyond 3D bounding boxes, encouraging the networks to learn more comprehensive 3D shape representations and potentially improving their ability to infer complete object geometry from sparse, partial observations.

\section{Conclusions}
We presented MVAT, a novel weakly supervised 3D object detection framework that addresses fundamental challenges in 3D perception using 2D bounding box image annotations only.
By aggregating object-specific point clouds across multiple frames, our approach overcomes the limitations of projection ambiguity and partial visibility inherent in single-frame methods.
The multi-view supervision strategy ensures consistency across camera perspectives, while our Teacher-Student knowledge transfer mechanism effectively generalizes from aggregated "easy" samples to isolated frame "hard" samples. \\

This work was supported by a French government grant managed by the Agence  Nationale de la Recherche under the France 2030 program with the reference ”ANR-23-
DEGR-0001”. This publication was made possible by the use of the FactoryIA supercomputer, financially supported by the Ile-De-France Regional Council.

{
    \small
    \bibliographystyle{ieeenat_fullname}
    \bibliography{main}
}

\clearpage
\setcounter{page}{1}
\maketitlesupplementary
\maketitle
\appendix

\section{Implementation Details}

\paragraph{Data Pre-processing}
For 2D mask generation from box prompts, we use SAM~2 and filter out masks with a confidence score below 0.6. When identifying static objects, we use a centroid variance threshold of 0.3m across frames. For the subsequent cleaning of aggregated point clouds ($P_{\text{agg}}^{s}$) before coarse box estimation, we apply DBSCAN with an $\epsilon$ of 0.5m and a minimum of 10 points per cluster.

\paragraph{Motion Classification}
To reliably separate static from moving instances for temporal aggregation, we implement a straightforward yet robust motion analysis procedure based on an object's displacement in a global reference frame.

First, leveraging the provided ego-pose data from the dataset, all object-centric point clouds within a given sequence are transformed into a common global coordinate frame. This step ensures that any observed motion is from the object itself, not due to the ego-vehicle's movement.

For a given object instance~$j$, which is tracked across a set of frames~$\mathcal{F}_{j}$, we compute its point cloud centroid~$c_{t,j}$ in this global frame for each timestamp~$t \in \mathcal{F}_{j}$. An object is then classified as \textbf{static} if the maximum Euclidean distance between any pair of its observed centroids throughout its entire track is below a predefined displacement threshold, $\tau_{\text{static}}$.

Formally, an object~$j$ is considered static if:
\begin{equation}
\max_{i,k \in \mathcal{F}_{j}} \|c_{i,j} - c_{k,j}\|_{2} < \tau_{\text{static}}
\end{equation}

In our experiments, we empirically set $\tau_{\text{static}} = 0.5$meter, a value that robustly accounts for potential minor centroid jitter caused by partial occlusions, point cloud sparsity, or slight inaccuracies in the provided ego-pose estimation. This criterion, which considers an object's full trajectory rather than just consecutive frames, effectively identifies globally stationary objects whose point clouds can be confidently aggregated.

\paragraph{3D Box Coarse Estimation with Geometric Verification}

The goal of this step is to derive a coarse-yet-robust 3D pseudo-box, $B_{3D}^{\text{coarse}}$, from each aggregated static point cloud cluster, $C^*$. We acknowledge that naively fitting a bounding box using methods like Principal Component Analysis (PCA) can be fragile. This can lead to significant errors, particularly for sparse point clouds or those with irregular distributions arising from occlusions (e.g., 'L'-shaped views). To mitigate this, our coarse estimation pipeline incorporates a geometric verification step to filter out unreliable instances before they are used to train the Teacher network.

The process is twofold:
\begin{enumerate}
    \item \textbf{Initial Box Hypothesis via PCA}: We first generate an initial box hypothesis. By applying PCA to the bird's-eye view (BEV) projection of the points in~$C^*$, we determine the primary orientation and initial extent of the object. This provides an estimate for the BEV center $(c_x, c_y)$, size~$(l, w)$, and heading~$\theta$. The vertical parameters ($c_z, h$) are subsequently derived from the min/max z-coordinates of the points.

    \item \textbf{Geometric Consistency Verification}: To validate the quality of this PCA-derived box, we perform a consistency check in the BEV plane. We compute the \textbf{2D convex hull} of the point cluster, which provides a tight geometric footprint of the object's observed shape. We then calculate the \textbf{Intersection-over-Union (IoU)} between the area of the PCA-estimated BEV box and the area of this convex hull. An instance is deemed geometrically reliable and retained for the Teacher's burn-in phase only if this shape-consistency IoU exceeds a predefined threshold, $\tau_{\text{IoU}}$:
    \begin{equation}
        \label{eq:iou_check}
        \text{IoU}(\text{Box}_{\text{PCA}}^{\text{BEV}}, \text{ConvexHull}(C^{*}_{\text{BEV}})) > \tau_{\text{IoU}}
    \end{equation}
\end{enumerate}

Instances failing this check are discarded from the initial training set for the Teacher. In our experiments, we set $\tau_{\text{IoU}} = 0.6$. This verification step is crucial as it prunes cases where PCA fails, ensuring that the Teacher is bootstrapped using only pseudo-labels with a high degree of geometric fidelity to the underlying point cloud data.

\paragraph{Pseudo-Label Filtering for Student Training}
A critical step before the knowledge distillation phase is to generate a high-quality set of pseudo-labels, $B_{3D}^{\text{Teacher}}$, to supervise the Student network. A naive transfer of all the Teacher's predictions would propagate errors and degrade the Student's performance. To prevent this, and to address any ambiguity in our filtering methodology, we employ a rigorous two-stage process on the Teacher's outputs for each object instance.

\begin{enumerate}
    \item \textbf{Class-Consistency Check}: First, we ensure semantic correctness. Our Teacher network outputs both a 3D box and a class prediction for each object. We perform an initial filtering pass by comparing the Teacher's predicted class with the ground-truth class label that is provided with the original 2D bounding box annotation. Any instance where the predicted class does not match the ground-truth class is immediately discarded. This step eliminates major classification failures and ensures the Student learns the correct class associations.

    \item \textbf{Confidence-Based Filtering}: Second, among the class-consistent predictions, we filter based on the model's certainty. The Teacher network also outputs a confidence score for each prediction, which reflects its certainty in both the classification and the localization quality of the 3D box. We only retain an instance for the Student's training set if its associated confidence score exceeds a predefined, class-specific threshold, $\tau_{\text{conf}}^{(c)}$, where $c$ denotes the object class.
\end{enumerate}

This two-stage verification ensures that the Student is trained on a "clean" dataset of pseudo-labels, where objects are correctly classified and their 3D boxes are predicted with high confidence by the Teacher. These thresholds are set based on the validation set to balance the trade-off between the quantity and quality of pseudo-labels. For example, in our experiments, we used values such as $\tau_{\text{conf}}^{(\text{Car})} = 0.5$ and $\tau_{\text{conf}}^{(\text{Pedestrian})} = 0.4$. This transparent and reproducible filtering mechanism is essential for effective knowledge distillation and prevents the propagation of low-quality predictions.

\paragraph{Training Details}
Our framework is implemented on top of the CenterPoint detector architecture.
For the \textit{Teacher Burn-in} phase, we train on the filtered set of static objects.
The loss weight for the training of the Teacher (Eq.~\ref{eq:Teacher_loss}) is set to $\lambda = 0.5$ for the 2D multi-view loss.
For the \textit{Student Distillation} phase, the teacher's weights are frozen and the student network is trained from scratch using the full dataset.
The loss weight for the student's training (Eq.~\ref{eq:Student_loss}) is set to $\gamma = 0.5$ for the 2D multi-view loss. 

\paragraph{PV-RCNN settings}
The PV-RCNN model parameters were primarily configured based on the KITTI dataset's optimal settings, which were then consistently applied across Waymo and nuScenes datasets. Training involved 80 epochs with a batch size of 24 and a learning rate of 0.01. The 3D voxel CNN backbone had four levels with feature dimensions 16, 32, 64, and 64 respectively. The Voxel Set Abstraction (VSA) module utilized two neighboring radii for each level, specifically (0.4m, 0.8m), (0.8m, 1.2m), (1.2m, 2.4m), and (2.4m, 4.8m), with raw points also using (0.4m, 0.8m). Keypoint sampling was set at 2,048. The Rol-grid pooling operation uniformly sampled $6 \times 6 \times 6$ grid points per 3D proposal, with two neighboring radii of (0.8m, 1.6m). Voxel sizes were (0.05m, 0.05m, 0.1m), and the detection range was $[0, 70.4]m$ (X), $[-40, 40]m$ (Y), and $[-3, 1]m$ (Z). The model was trained end-to-end using the ADAM optimizer, incorporating random flipping along the X axis, global scaling with a random factor from $[0.95, 1.05]$, global rotation around the Z axis with a random angle from $[-\frac{\pi}{4}, \frac{\pi}{4}]$, and ground-truth sampling for data augmentation. During inference, the top-100 proposals generated by the 3D voxel CNN were kept with a 3D IoU threshold of 0.7 for non-maximum-suppression (NMS), and a final NMS threshold of 0.01 was applied. 

\textbf{The overall 3D training loss ($\mathcal{L}_{\text{3D}}$)} was a sum of the region proposal loss, 3D segmentation loss, and proposal refinement loss, all with equal weights. The region proposal loss includes focal loss for classification and smooth-L1 for anchor box regression. The 3D segmentation loss also uses focal loss for keypoint segmentation. The proposal refinement loss comprises an IoU-guided confidence prediction loss, which uses cross-entropy on a normalized IoU target, and a box refinement loss based on smooth-L1.

\paragraph{CenterPoint settings}
We used the exact same architecture as in the original paper for the CenterPoint model. The architecture consists of a standard 3D backbone that extracts map-view feature representation from Lidar point-clouds. Subsequently, a 2D CNN architecture detection head identifies object centers and regresses to full 3D bounding boxes using center features. This box prediction is then utilized to extract point features at the 3D centers of each face of the estimated 3D bounding box, which are passed into an MLP to predict an IoU-guided confidence score and box regression refinement. All first-stage outputs share a $3 \times 3$ convolutional layer, Batch Normalization, and ReLU, with each output then branching into two $3 \times 3$ convolutions separated by a batch norm and ReLU. The second stage employs a shared two-layer MLP, including a batch norm, ReLU, and Dropout with a drop rate of 0.3, followed by two branches of three fully-connected layers for confidence score and box regression prediction.

The CenterPoint model utilizes specific hyperparameters for optimal performance across different datasets, namely nuScenes and Waymo Open Dataset. For the nuScenes dataset, the model is optimized using the AdamW optimizer with a one-cycle learning rate policy, setting the maximum learning rate at $10^{-3}$, a weight decay of 0.01, and momentum ranging from 0.85 to 0.95. Training is conducted with a batch size of 16 over 20 epochs on 4 V100 GPUs. The detection range for nuScenes is defined as [-51.2m, 51.2m] for the X and Y axes and [-5m, 3m] for the Z axis. Depending on the encoder, the voxel size for CenterPoint-Voxel is (0.1m, 0.1m, 0.2m) and the grid size for CenterPoint-Pillars is (0.2m, 0.2m). Data augmentation techniques include random flipping along both X and Y axes , global scaling with a random factor between [0.95, 1.05], random global rotation within $[-\pi/8, \pi/8]$ , and ground-truth sampling to address class distribution imbalances. For the second stage refinement during training, 128 boxes are randomly sampled with a 1:1 positive-negative ratio, where a positive proposal overlaps with a ground truth annotation by at least 0.55 IoU. During inference, the second stage processes the top 500 predictions after Non-Maxima Suppression (NMS). For the nuScenes test set submission, a finer input grid size of $0.075m \times 0.075m$ was used, incorporating two separate deformable convolution layers in the detection head.

Conversely, for the Waymo Open Dataset, the CenterPoint model employs a learning rate of $3 \times 10^{-3}$ and is trained for 30 epochs. The detection range is set to [-75.2m, 75.2m) for the X and Y axes, and [-2m, 4m] for the Z axis. CenterPoint-Voxel utilizes a voxel size of (0.1m, 0.1m, 0.15m), while CenterPoint-Pillar uses a grid size of (0.32m, 0.32m). Data augmentation for Waymo includes random flipping along both X and Y axes, global scaling between [0.95, 1.05], and a random global rotation of $[-\pi/4, \pi/4]$. For various ablation studies, the model was finetuned for 6 epochs with the second stage refinement modules.

\paragraph{Dataset Considerations}
Our experiments are conducted on sequential datasets like nuScenes and Waymo. While the KITTI dataset is prominent in 3D detection, it does not provide the sequential, multi-view data that is essential for our method's temporal aggregation strategy.

\section{Number of view frames per object for each class.}

\begin{figure}[tb]
    \centering
    \includegraphics[width=1\linewidth]{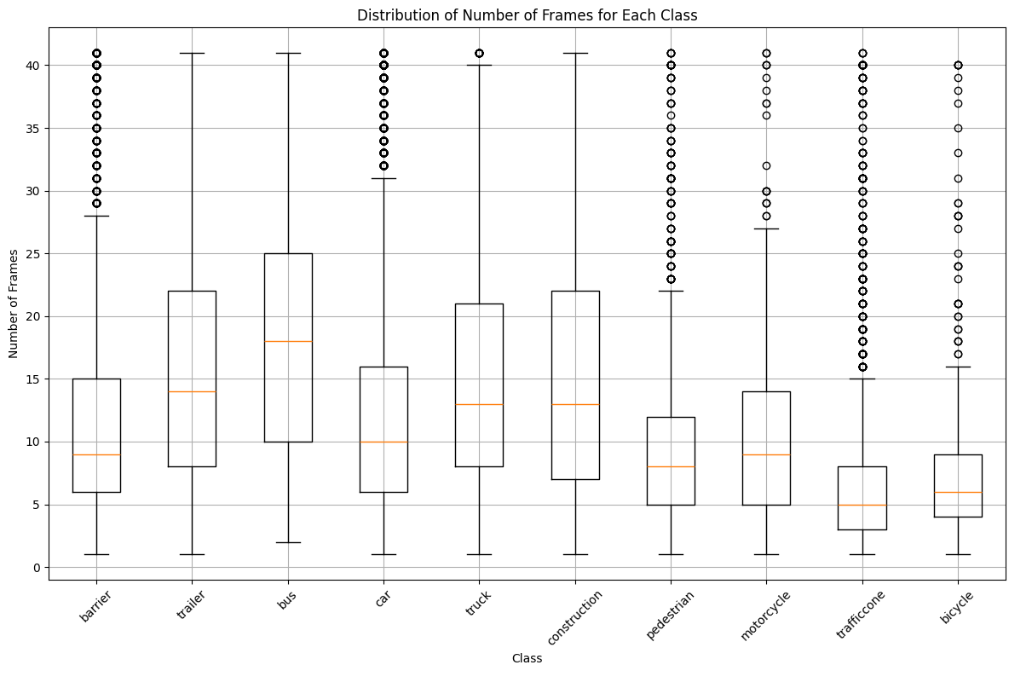}
    \caption{Distribution of the number of frames for objects of each class}
    \label{fig:frames}
\end{figure}

To better understand the characteristics of our dataset and the potential benefits of temporal aggregation, we analyze the distribution of view frames per object across different classes. \autoref{fig:frames} illustrates this distribution, revealing significant variations among object categories. The median number of frames in which an object appears varies substantially across classes. Traffic cones exhibit the lowest median visibility at only 5 frames per instance, which can be attributed to their small size and frequent occlusion in urban environments. In contrast, buses maintain the highest visibility with a median of 18 frames per instance, likely due to their large size and tendency to remain in the field of view for extended periods.
This class-dependent distribution of temporal visibility has important implications for our method. Classes with higher frame counts (buses, trucks, and cars) benefit more from our temporal aggregation approach, as more views contribute to a more complete 3D representation. Conversely, objects with fewer frames (traffic cones, pedestrians) present a greater challenge, as the temporal aggregation provides more limited improvements over single-frame detection.

\section{Qualitative results of the PCA BEV box estimation.}

\begin{figure*}[ht]
    \centering
    \includegraphics[width=0.5\linewidth]{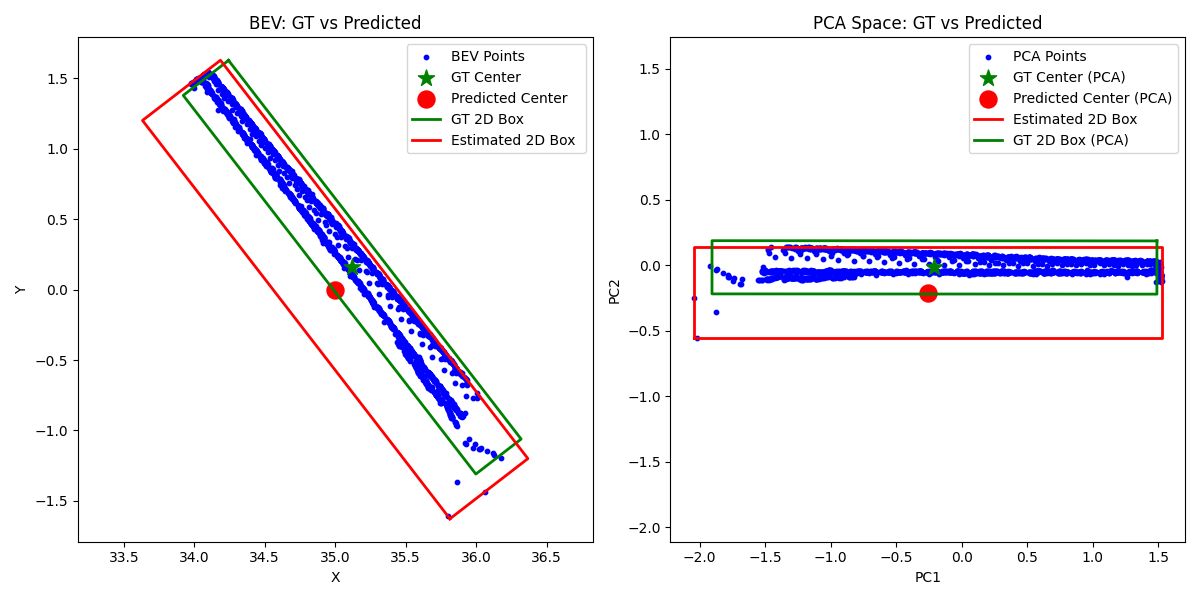}
    \caption{Barrier Example}
    \includegraphics[width=0.5\linewidth]{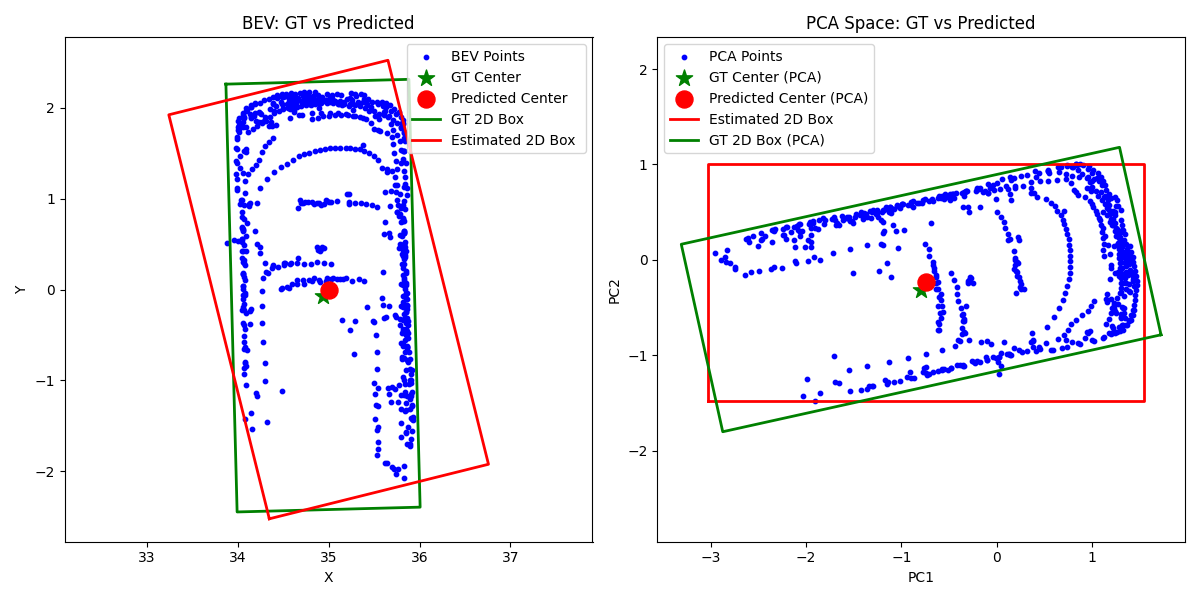}
    \caption{Car Example 1}
    \includegraphics[width=0.5\linewidth]{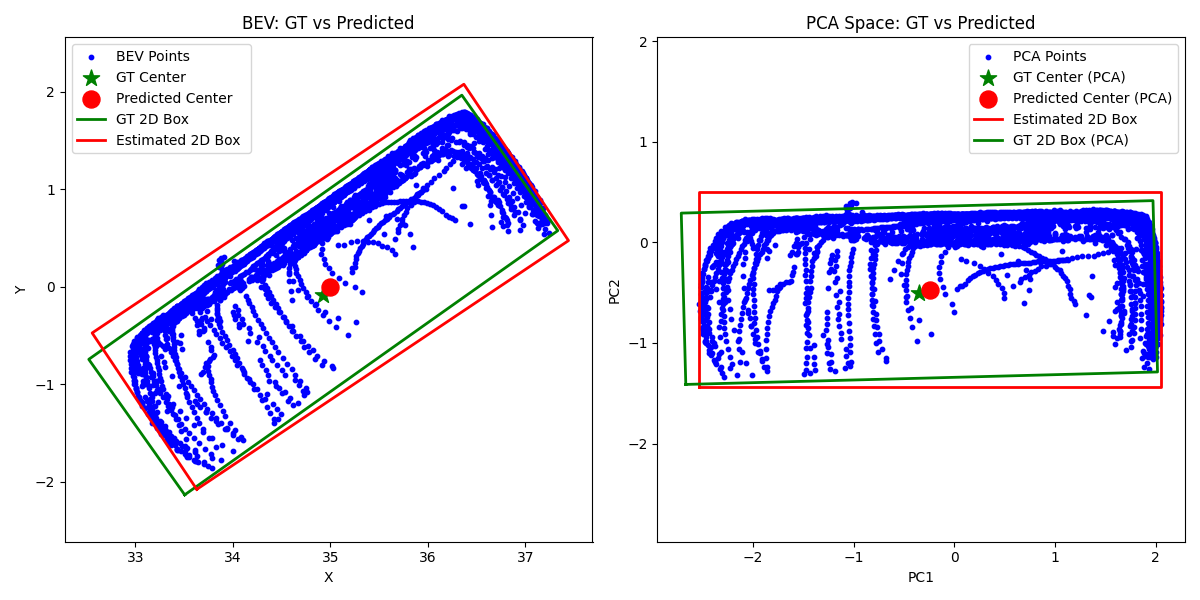}
    \caption{Car Example 2}
    \includegraphics[width=0.5\linewidth]{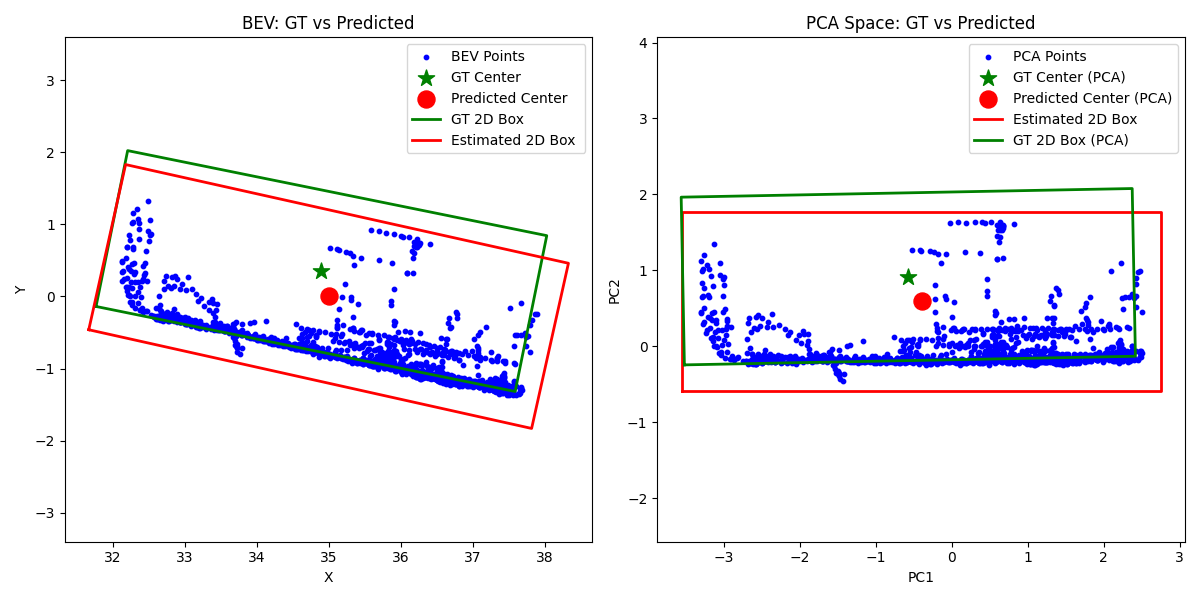}
    \caption{Truck Example}
    \label{fig:pca}
\end{figure*}

We present visualizations of our coarse 3D box estimation, where PCA is applied to the BEV-projected aggregated point clouds. As shown in \autoref{fig:pca}, when objects are clearly visible, our method accurately estimates the 3D center, dimensions, and orientation. For example, the Barrier, Car, and Truck examples illustrate that the estimated boxes closely align with the object boundaries and correctly capture their orientations. These results demonstrate that our PCA-based estimation reliably produces high-quality coarse 3D annotations under favorable conditions, providing robust supervisory signals for subsequent 3D detection stages.

\section{Detailed quantitative results of the coarse 3D box estimation.}

\begin{table*}[htbp]
\centering
 \begin{adjustbox}{width=\linewidth}
\begin{tabular}{cccccccccccc}
\hline
\textbf{Category} & \textbf{Barrier} & \textbf{Car} & \textbf{Truck} & \textbf{Motorcycle} & \textbf{Traffic Cone} & \textbf{Pedestrian} & \textbf{Construction} & \textbf{Bicycle} & \textbf{Bus} & \textbf{Trailer} & \textbf{Average} \\
\hline
\textbf{3D IoU} & 44.7 & 53.3 & 49.5 & 50.8 & 48.6 & 53.6 & 49.4 & 46.2 & 42.7 & 53.2 & \textbf{49.2} \\

\hline
\end{tabular}
 \end{adjustbox}
\caption{Evaluation in 3D IoU between the 3D coarse estimations \(B_{\text{3D}}^{\text{coarse}}\) and ground truth 3D boxes for the samples used to train the teacher network.}
\label{tab:iou_all}
\end{table*}

\begin{figure*}[tb]
    \centering
    \includegraphics[width=\linewidth]{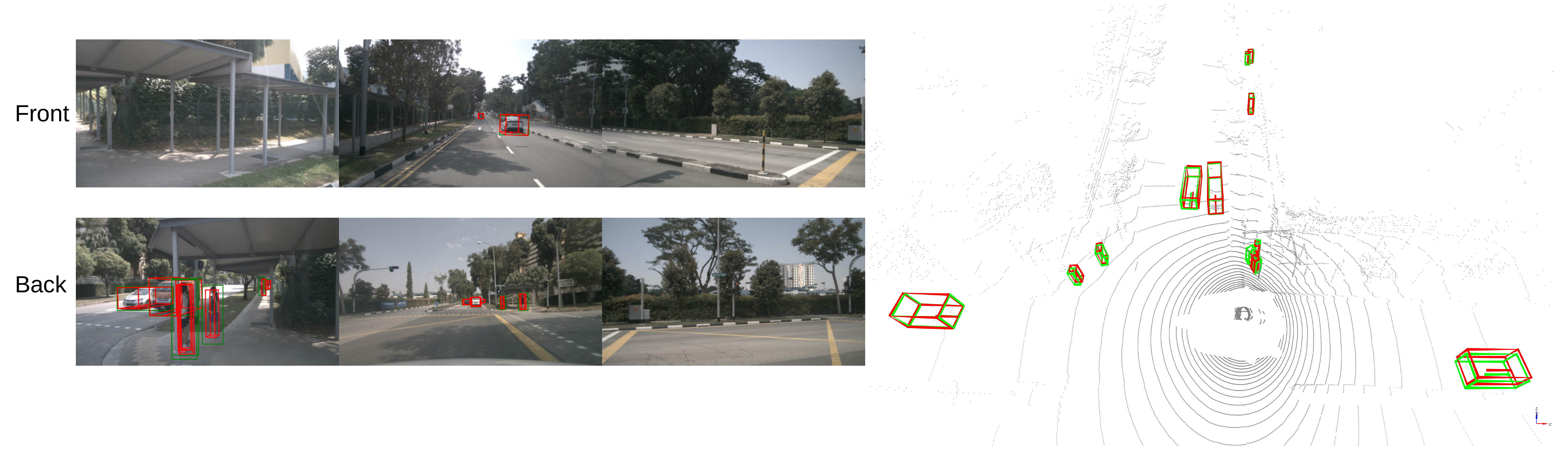}

    \includegraphics[width=\linewidth]{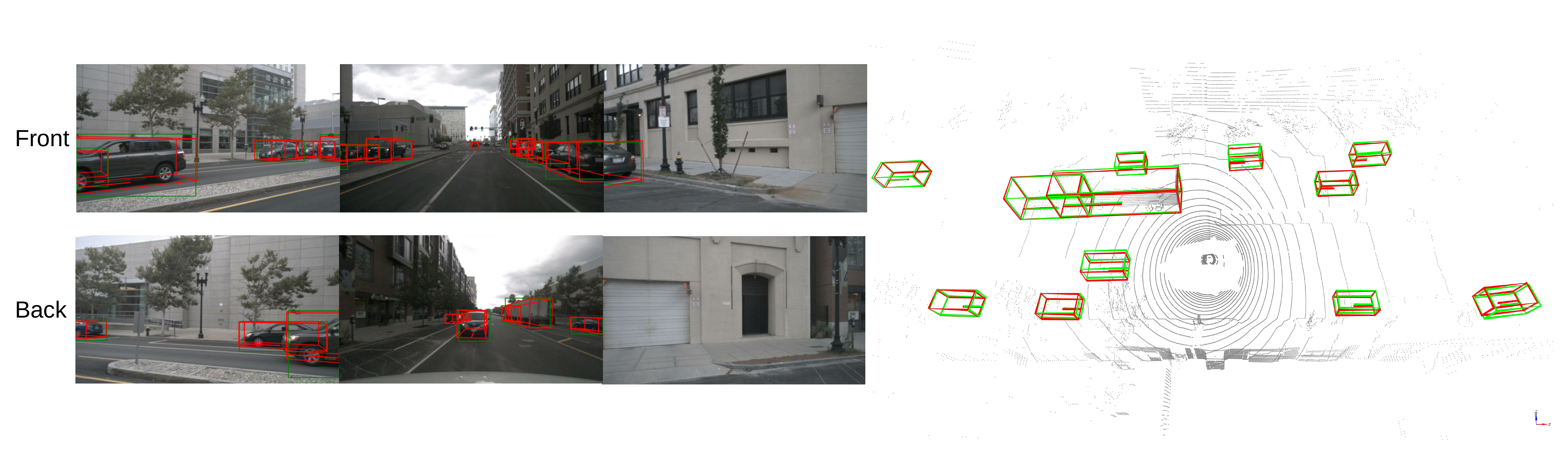}
    \caption{Qualitative results of CenterPoint~\cite{yin2021center} trained using 3D pseudo-labeled by MVAT on the nuScenes validation set. We show the ground truth boxes in green and the predictions in red.}
    \label{fig:visusupp2}
\end{figure*}


\end{document}